\def\year{2022}\relax
\documentclass[letterpaper]{article} 
\usepackage{aaai22}  
\usepackage{times}  
\usepackage{helvet} 
\usepackage{courier}  
\usepackage[hyphens]{url}  
\usepackage{graphicx} 
\usepackage{natbib}  
\usepackage{caption} 
\usepackage{amsmath}
\usepackage{amssymb}
\usepackage{algorithm}
\usepackage{algorithmic}
\usepackage{setspace}
\usepackage{subfig}
\usepackage{adjustbox}
\usepackage{enumitem}

\usepackage{tikz}
\usetikzlibrary{arrows,automata,positioning}
\usetikzlibrary{shapes.multipart}
\usetikzlibrary{decorations.markings}
\usetikzlibrary{decorations.pathreplacing}

\newcommand{\tS}{\text{t}}   
\newcommand{\tSi}{\tau}   

\newcommand{\cB}{\mathcal{B}}
\newcommand{\cC}{\mathcal{C}}
\newcommand{\cD}{\mathcal{D}}
\newcommand{\cE}{\mathcal{E}}
\newcommand{\cJ}{\mathcal{J}}
\newcommand{\cL}{\mathcal{L}}

\newcommand{\cP}{\mathcal{P}}
\newcommand{\cR}{\mathcal{R}}
\newcommand{\cS}{\mathcal{S}}
\newcommand{\cT}{\mathcal{T}}
\newcommand{\cU}{\mathcal{U}}
\newcommand{\cV}{\mathcal{V}}
\newcommand{\cX}{\mathcal{X}}

\newcommand{\EEc}[2]{\mathbb{E}\left[#1\;\middle\lvert\;#2\right]}

\newcommand{\diag}{\text{diag}}
\newcommand{\real}{\mathbb{R}}

\newtheorem{theorem}{Theorem}[section]
\newtheorem{lemma}[theorem]{Lemma}
\newtheorem{definition}[theorem]{Definition}
\newenvironment{proof}[1][Proof]{\begin{trivlist}
\item[\hskip \labelsep {\bfseries #1}]}{\end{trivlist}}

\title{Globally Optimal Hierarchical Reinforcement Learning for Linearly-Solvable Markov Decision Processes}
\author{
    Guillermo Infante, 
    Anders Jonsson, 
    Vicen\c{c} G\'omez 
}
\affiliations{
    Dept.~Information and Communication Technologies, 
  Universitat Pompeu Fabra,
  Barcelona (Spain)
    \{guillermo.infante,anders.jonsson,vicen.gomez\}@upf.edu

}

\begin{document}

\maketitle

\begin{abstract}
We present a novel approach to hierarchical reinforcement learning for linearly-solvable Markov decision processes. Our approach assumes that the state space is partitioned, and defines subtasks for moving between the partitions. We represent value functions on several levels of abstraction, and use the compositionality of subtasks to estimate the optimal values of the states in each partition. The policy is implicitly defined on these optimal value estimates, rather than being decomposed among the subtasks. As a consequence, our approach can learn the globally optimal policy, and does not suffer from non-stationarities induced by high-level decisions. If several partitions have equivalent dynamics, the subtasks of those partitions can be shared.
We show that our approach is significantly more sample efficient than that of a flat learner and similar hierarchical approaches when the set of boundary states is smaller than the entire state space.
\end{abstract}

\section*{Introduction}
A major challenge in reinforcement learning is to design agents that are able to learn efficiently and to adapt their existing knowledge to solve new tasks.

One way to reduce the complexity of learning is hierarchical reinforcement learning~\citep{sutton1999between,dietterich2000hierarchical,bartomahadevan}. By decomposing a task into subtasks, each of which can be solved independently, a solution to the original task can then be composed of the solutions to the subtasks. If each subtask is easier to solve than the original task, this may significantly reduce the learning effort of an agent that is learning to perform the task.

We consider Linearly-solvable Markov decision processes (LMDPs), a class of control problems 
whose Bellman optimality equations are linear in the (exponentiated) value function~\citep{Kappen2005,TodorovNIPS2007}. Because of this, solution methods for LMDPs are more efficient than those for general Markov decision processes (MDPs). 
Though not as expressive as MDPs, LMDPs can nevertheless model a wide range of decision problems, and there exist methods for approximating MDPs with LMDPs~\citep{TodorovNIPS2007}.

LMDPs frequently appear under the names of path-integral or Kullback-Leibler control in the context of optimal control as probabilistic inference~\citep{KappenML2012,djbook,kappenbook}. LMDPs are also strongly related to maximum-entropy reinforcement learning, which is known to have favorable properties and is quickly becoming the state-of-the-art for reinforcement learning~\citep{ziebart,pmlr-v48-mniha16,pmlr-v80-haarnoja18b,levine2018reinforcement,NEURIPS2020_2c6a0bae,pmlr-v130-bas-serrano21a}.



One of the computational advantages of LMDPs is compositionality, which allows for zero-shot learning of new skills by linearly combining previously learned base skills which only differ in their cost or reward at boundary states~\cite{TodorovNIPS2009,animation}.


In this paper we propose a novel approach to hierarchical reinforcement learning in LMDPs that takes advantage of the compositionality of LMDPs. Our approach assumes that the state space is partitioned into subsets, and the subtasks consist in moving between these partitions. The subtasks are parameterized on the current value estimates of boundary states. Instead of solving the subtasks each time the value estimates change, we take advantage of compositionality to express the solution to an arbitrary subtask as a linear combination of a set of base LMDPs. The result is a form of value function decomposition which allows us to express an estimate of the optimal value of an arbitrary state as a combination of multiple value functions with smaller domains.

Concretely, our work makes the following contributions:
\begin{itemize}
\item We define a novel scheme based on compositionality for solving subtasks, defining local rewards that constitute a convenient basis for composite rewards.
\item The subtask decomposition is at the level of the value function, not of the actual policy. Hence our approach does not suffer from non-stationarity in the online setting, unlike approaches that select among subtasks whose associated policies are being learned.
\item Even though the subtasks have local reward functions, under mild assumptions our approach converges to the globally optimal value function.
\item We analyze experimentally our proposed learning algorithm and show in two classical domains 
that it is more sample efficient compared to a flat learner and similar hierarchical approaches when the set of boundary states is smaller than the entire state space.
\end{itemize}

\section*{Related Work}
Several authors have recently exploited concurrent compositionality of tasks in the context of transfer learning.~\citet{van2019composing} use the linear compositionality of LMDPs to solve new tasks that can be expressed as combinations of a series of existing base tasks. They show that, while disjunctions of base tasks (OR-compositionality) can be performed exactly, the AND composition (when the goals of base tasks partially overlap) can only be performed approximately.

\citet{haarnoja2018composable} exploit a similar idea to transfer knowledge from existing tasks to new tasks by averaging their reward functions.~\citet{hunt2019composing} further extended this by introducing the so-called compositional optimism, and apply divergence correction in case compositionality does not transfer well.

More recently, \citet{algebra} derive a formal characterization of union and intersection of tasks in terms of Boolean algebra. They show that learning (extended) value functions that account for all achievable goals, exact zero-shot transfer learning using both AND- and OR- compositionality is possible, achieving an exponential increase in skills compared to the previous works.

All the aforementioned results are derived for general MDPs with \emph{deterministic} dynamics and, possibly, entropy regularization. This setting is no more general than the class of LMDPs or path-integral control.



In this work, we aim to integrate both concurrent task composition, as done in the above approaches, together with hierarchical composition, where skills are chained in a temporal sequence, under the framework of LMDPs.

Several authors have proposed hierarchical versions of LMDPs.~\citet{conf/icaps/Jonsson16} extend MAXQ~\citep{dietterich2000hierarchical} to LMDPs by defining subtasks that represent high-level decisions. The top-level policy chooses multi-step transitions, which introduces non-stationarity in the high-level decision process if subtasks are learned concurrently, and also prevents global optimality. The authors discuss the idea of compositionality, but do not explore the concept further.~\citet{saxe2017hierarchy} propose a hierarchical multi-task architecture that does exploit compositionality.
Their Multitask LMDP maintains a parallel distributed representation of tasks, reducing the complexity through stacking. However, the approach requires to augment the state space with many additional boundary (subtask) states. Further, the stacking introduces additional costs (cf. their Equation 10), and does not provide global optimality.

The Options Keyboard \citep{optionkeyb} combines a successor feature representation with generalized policy improvement to obtain subtask policies from a set of base subtasks without learning, similar to our use of subtask compositionality. However, unlike in our approach, the composition weights have to be set manually, and although the composed policy is guaranteed to be better than the individual base policies, it is not guaranteed to be optimal.


Our work is similar to that of \citet{conf/nips/Wen20} in that we define a hierarchical decomposition based on a partition of the state space, and exploit the equivalence of subtasks to reduce the learning effort. Unlike previous work, however, our approach is not restricted to single initial states, does not suffer from non-stationarity in the online setting, proposes a more general definition of equivalence that captures more structure, and guarantees convergence to the optimal value function for stochastic dynamics.

The concept of equivalent subtasks is strongly related to factored (L)MDPs, which capture conditional independence among a set of state variables~\cite{boutilier95,DBLP:conf/uai/KollerP00}. Equivalence arises whenever a subset of state variables are conditionally independent of another subset. Several authors have shown how to automatically discover the structure of factored MDPs from experience~\cite{DBLP:conf/aaai/StrehlDL07,Kolobov2012}, which in turn could be used to define equivalence classes of subtasks.

\section*{Background}


Given a finite set $\cX$, let $\Delta(\cX)=\{p\in\real^\cX:\sum_x p(x)=1,\linebreak p(x)\geq 0\;(\forall x)\}$ denote the probability simplex on $\cX$. Given a probability distribution $p\in\Delta(\cX)$, let $\cB(p)=\{x\in\cX:p(x) > 0\}\subseteq\cX$ denote the support of $p$.

\subsection{Linearly-Solvable Markov Decision Processes}

A linearly-solvable Markov decision process, or LMDP~\citep{Kappen2005,TodorovNIPS2007}, can be defined as a tuple $\cL=\langle\cS,\cT,\cP,\cR,\cJ\rangle$, where $\cS$ is a set of non-terminal states, $\cT$ is a set of terminal states, $\cP:\cS\rightarrow\Delta(\cS^+)$ is an uncontrolled transition function, $\cR:\cS\rightarrow\real$ is a reward function for non-terminal states, and $\cJ:\cT\rightarrow\real$ is a reward function for terminal states. We use $\cS^+=\cS\cup\cT$ to denote the full set of states, and $S^+=|\cS^+|$ (resp.~$S=|\cS|$) to denote the number of (non-terminal) states. 
We also use $B=\max_{s\in\cS}|\cB(\cP(\cdot|s))|$ to denote an upper bound on the support of $P$.

The learning agent follows a policy $\pi:\cS\rightarrow\Delta(\cS^+)$ that, for each non-terminal state $s\in\cS$, chooses a probability distribution over next states in the support of $\cP(\cdot|s)$, i.e.~$\pi(\cdot|s)\in\Delta(\cB(\cP(\cdot|s))$. In each round $t$, the learning agent observes a state $s_t\in\cS^+$. If $s_t$ is non-terminal, the agent transitions to a new state $s_{t+1}\sim\pi(\cdot|s_t)$ and receives an immediate reward
\[
\cR(s_t,\pi) = \cR(s_t) - \lambda\cdot\mathrm{KL}(\pi(\cdot|s_t)\Vert\, \cP(\cdot|s_t)),
\]
where $\cR(s_t)$ is the reward associated with state $s_t$, $\mathrm{KL}(\pi(\cdot|s_t)\Vert\, \cP(\cdot|s_t))$ is the Kullback-Leibler divergence between $\pi(\cdot|s_t)$ and $\cP(\cdot|s_t)$, and $\lambda$ is a temperature parameter. Hence the agent can set the probability distribution $\pi(\cdot|s_t)$ freely, but gets penalized for deviating from the uncontrolled distribution $\cP(\cdot|s_t)$. On the other hand, if $s_t$ is terminal, the agent receives reward $\cJ(s_t)$ and then the current episode ends. The aim of the agent is to compute a policy $\pi$ that maximizes the expected future reward (i.e.~value), defined in each non-terminal state $s\in\cS$ as
\[
v^\pi(s) = \EEc{\sum_{t=1}^{T-1} \cR(S_t,\pi) + \cJ(S_T)}{S_1 = s}.
\]
Here, $T$ is a random variable representing the time at which the current episode ends, and $S_t$ is a random variable representing the state at time $t$. The expectation is over the stochastic choice of next state $S_{t+1}\sim\pi(\cdot|S_t)$ at each time~$t$, and the time $T$ it takes for the episode to end. We assume that the reward of all non-terminal states is negative, i.e.~$\cR(s)<0$ for each $s\in\cS$. As a consequence, $\cR(s,\pi)<0$ holds for any policy $\pi$, and the value $v^\pi(s)$ has a well-defined upper bound. 

We are interested in computing the optimal value function $v^*:\cS\rightarrow\real$, i.e.~the maximum expected future reward among all policies. For simplicity, in what follows we omit the asterisks and refer to the optimal value function simply as the value function. We extend the value function to each terminal state $\tS\in\cT$ by defining $v(\tS)\equiv\cJ(t)$. The value function $v$ satisfies the Bellman equations
\begin{align*}
& \frac 1 \lambda v(s) = \frac 1 \lambda \max_\pi \left[ \cR(s,\pi) + \mathbb{E}_{s'\sim\pi(\cdot|s)} v(s') \right] \\
 &= \frac{1}{\lambda} \cR(s) + \max_\pi \mathbb{E}_{s'\sim\pi(\cdot|s)} \left[ \frac 1 \lambda v(s') - \log \frac {\pi(s'|s)} {\cP(s'|s)} \right] \;\; \forall s.
\end{align*}
We introduce the notation $z(s)=e^{v(s)/\lambda}$ for each $s\in\cS^+$, and often abuse notation by referring to $z(s)$ as the (optimal) value of $s$. The maximization in the Bellman equations can be resolved analytically, yielding the following Bellman equations that are linear in $z$:
\begin{equation}\label{eq:z}
z(s) = e^{\cR(s)/\lambda} \sum_{s'}\cP(s'|s)z(s').
\end{equation}
We can express the Bellman equation in matrix form by defining an $S\times S$ diagonal reward matrix $R=\diag(e^{\cR(\cdot)/\lambda})$ and an $S\times S^+$ stochastic transition matrix $P$ whose entries $(s,s')$ equal $\cP(s'|s)$. We also define a vector $\bf z$ that stores the values $z(s)$ for each non-terminal state $s\in\cS$, and a vector $\bf z^+$ extended to all states in $\cS^+$. We can now write the Bellman equations in matrix form:
\begin{equation}\label{eq:matrixz}
{\bf z} = R P {\bf z^+}.
\end{equation}
Given $z$, the optimal policy $\pi$ is given by the following expression for each pair of states $(s,s')$:
\begin{equation}\label{eq:pi}
\pi(s'|s) = \frac {\cP(s'|s)z(s')} {\sum_{s''} \cP(s''|s)z(s'')}.
\end{equation}

The solution for $z$ corresponds to the largest eigenvector of $RP$.
If the dynamics $\cP$ and $\cR$ are known, we can iterate~\eqref{eq:matrixz}~\citep{TodorovNIPS2007}.
Alternatively, we can incrementally learn an estimate $\hat{z}$ using stochastic updates based on state transitions sampled from the uncontrolled dynamics $(s_t,r_t,s_{t+1})$
\[
\hat{z}(s_t) \leftarrow (1-\alpha_t)\hat{z}(s_t) + \alpha_t e^{r_t/\lambda}\hat{z}(s_{t+1}),
\]
where $\alpha_t$ is a learning rate. 
The above update rule is called Z-learning~\citep{TodorovNIPS2007} and suffers from slow convergence in very large state spaces and when the optimal policy differs substantially from the uncontrolled dynamics $\cP$.
A better choice is importance sampling, which uses samples from the estimated policy $\hat{\pi}$ derived from the estimated values $\hat{z}$ and \eqref{eq:pi} and updates $\hat{z}$ according to the following update
\begin{align}\label{eqn:zlearning-imp}
\hat{z}(s_t) \leftarrow (1-\alpha_t)& \hat{z}(s_t) + \alpha_t e^{r_t/\lambda}\hat{z}(s_{t+1})\frac {\cP(s_{t+1}|s_t)} {\hat{\pi}(s_{t+1}|s_t)}.
\end{align}
However, this requires local knowledge of $\cP(\cdot|s_t)$ to correct for the different sampling distribution.
Though this seems like a strong assumption, in practice $\cP$ usually has a simple form, e.g.~a random walk.
Further, as shown in \citet{conf/icaps/Jonsson16}, the corrected update rule in \eqref{eqn:zlearning-imp} can also be used to perform off-policy updates in case transitions are sampled using a policy different from $\hat{\pi}$,

\subsection{Compositionality}

\citet{TodorovNIPS2009} introduced the concept of compositionality for LMDPs. Consider a set of LMDPs $\{\cL_1,\ldots,\cL_n\}$, where each LMDP $\cL_i=\langle\cS,\cT,\cP,\cR,\cJ_i\rangle$ has the same components $\cS,\cT,\cP,\cR$ and only differ in the reward $\cJ_i(\tS)$ of each terminal state $\tS\in\cT$, as well as its exponentiated value $z_i(\tS)=e^{\cJ_i(\tS)/\lambda}$.

Now consider a new LMDP $\cL=\langle\cS,\cT,\cP,\cR,\cJ\rangle$ with the same components as the $n$ LMDPs above, except for $\cJ$. Assume that there exist weights $w_1,\ldots,w_n$ such that the exponentiated value of each terminal state $\tS\in\cT$ can be written as
\[
e^{\cJ(\tS)/\lambda} = z(\tS) = w_1z_1(\tS) + \ldots + w_nz_n(\tS) = \sum_{k=1}^n w_kz_k(\tS).
\]
Since the Bellman optimality equation of each non-terminal state $s\in\cS$ is linear in $z$, the optimal value of $s$ satisfies the same equation:
\[
z(s) = \sum_{k=1}^n w_kz_k(s).
\]
Consequently, if we previously compute the optimal values $z_1,\ldots,z_n$ of the $n$ LMDPs and know the weights $w_1,\ldots,w_n$, we immediately obtain the optimal values of the new LMDP $\cL$ without learning.


\section*{Hierarchical LMDPs}

In this section we describe our novel approach to hierarchical LMDPs. We first describe the particular form of hierarchical decomposition that we consider, and then present algorithms for solving a decomposed LMDP.

\subsection{Hierarchical Decomposition}

Our hierarchical decomposition is similar to that of~\citet{conf/nips/Wen20}. Formally, given an LMDP $\cL=\langle\cS,\cT,\cP,\cR,\cJ\rangle$, the set of non-terminal states $\cS$ is partitioned into $L$ subsets $\{\cS_i\}_{i=1}^L$. For each such subset $\cS_i$, we define an induced subtask $\cL_i=\langle\cS_i,\cT_i,\cP_i,\cR_i,\cJ_i\rangle$, i.e.~an LMDP whose components are defined as follows:
\begin{itemize}
\item The set of non-terminal states is $\cS_i$.
\item The set of terminal states $\cT_i=\{\tSi \in\cS^+\setminus\cS_i:\exists s\in \cS_i \; \text{s.t.} \; \tSi \in \cB(\cP(\cdot|s))\}$ includes all states in $\cS^+\setminus\cS_i$ (terminal or non-terminal) that are reachable in one step from a state in $\cS_i$.
\item $\cP_i:\cS_i\rightarrow\Delta(\cS_i^+)$ and $\cR_i:\cS_i\rightarrow\real$ are the restrictions of $\cP$ and $\cR$ to $\cS_i$, where $\cS_i^+=\cS_i\cup\cT_i$ denotes the full set of subtask states.
\item The reward of a terminal state $\tSi \in\cT_i$ equals $\cJ_i(\tSi)=\cJ(\tSi)$ if $\tSi\in\cT$, and $\cJ_i(\tSi)=\hat{v}(\tSi)$ otherwise, where $\hat{v}(\tSi)$ is the estimated value in $\cL$ of the non-terminal state in  $\tSi\in\cS \setminus \cS_i$.
\end{itemize}

Intuitively, if the reward $\cJ_i(\tSi)$ of each terminal state $\tSi\in\cT_i$ equals its optimal value $v(\tSi)$ for the original LMDP~$\cL$, then solving the subtask $\cL_i$ yields the optimal values of the states in $\cS_i$.
In practice, however, we only have access to an estimate $\hat{v}(\tSi)$ of the optimal value.
In this case, the subtask $\cL_i$ is {\em parameterized} on the value estimate $\hat{v}$ of terminal states in $\cT_i$, and each time the value estimate changes, we can solve $\cL_i$ to obtain a new value estimate
$\hat{v}(s)$ for each state $s\in\cS_i$.

We define a set of {\em exit states} $\cE=\cup_{i=1}^L\cT_i$, i.e.~the union of the terminal states of each subtask in $\{\cL_1,\ldots,\cL_L\}$. For convenience, we use $\cE_i=\cE\cap\cS_i$ to denote the set of (non-terminal) exit states in the subtask $\cL_i$. We also introduce the notation $K=\max_{i=1}^L|\cS_i|$, $N=\max_{i=1}^L|\cT_i|$ and $E=|\cE|$.

Just like \citet{conf/nips/Wen20}, we define a notion of equivalent subtasks.
\begin{definition}
Two subtasks $\cL_i$ and $\cL_j$ are equivalent if there exists a bijection $f:\cS_i\rightarrow\cS_j$ such that the transition probabilities and rewards of non-terminal states are equivalent through $f$.
\end{definition}
Unlike \citet{conf/nips/Wen20}, we do {\em not} require the sets of terminal states $\cT_i$ and $\cT_j$ to be equivalent. Instead, for each class of equivalent subtasks, our approach is to define a single subtask whose set of terminal states is the {\em union} of the sets of terminal states of subtasks in the class.

Formally, we define a set of equivalence classes $\cC=\{\cC_1,\ldots,\cC_C\}$, $C\leq L$, i.e.~a partition of the set of subtasks $\{\cL_1,\ldots,\cL_L\}$ such that all subtasks in a given partition are equivalent. We represent a single subtask $\cL_j=\langle\cS_j,\cT_j,\cP_j,\cR_j,\cJ_j\rangle$ per equivalence class $\cC_j\in\cC$. The components $\cS_j,\cP_j,\cR_j$ are shared by all subtasks in the equivalence class, while the set of terminal states is $\cT_j=\bigcup_{\cL_i\in\cC_j} \cT_i$, where the union is taken w.r.t. the bijection $f$ relating all equivalent subtasks. As before, the reward $\cJ_j$ of terminal states is parameterized on a given value estimate $\hat{v}$. We assume that each non-terminal state $s\in\cS$ can be easily mapped to its subtask $\cL_i$ and equivalence class $\cC_j$.

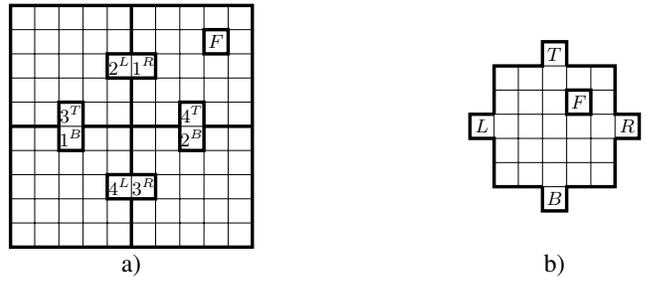
\begin{figure}[!t]
\begin{center}
\begin{adjustbox}{width=\columnwidth}
\begin{tikzpicture}
\draw[step=0.4,thin,shift={(0.2,0.2)}] (0.8,0.8) grid (4.8,4.8);
\draw[ultra thick] (1,1) rectangle (5,5);
\draw[ultra thick] (3,1) -- (3,1.8);
\draw[ultra thick] (3,2.2) -- (3,3.8);
\draw[ultra thick] (3,4.2) -- (3,5);
\draw[ultra thick] (1,3) -- (1.8,3);
\draw[ultra thick] (2.2,3) -- (3.8,3);
\draw[ultra thick] (4.2,3) -- (5,3);


\draw[ultra thick] (4.2,4.2) rectangle (4.6,4.6);
\draw[ultra thick] (3.8,2.6) rectangle (4.2,3.4);
\draw[ultra thick] (1.8,2.6) rectangle (2.2,3.4);
\draw[ultra thick] (2.6,3.8) rectangle (3.4,4.2);
\draw[ultra thick] (2.6,1.8) rectangle (3.4,2.2);

\node at (4.4,4.4) {\small $F$};
\node at (2,3.2) {\small $3^T$};
\node at (2,2.8) {\small $1^B$};
\node at (4,3.2) {\small $4^T$};
\node at (4,2.8) {\small $2^B$};
\node at (2.8,4) {\small $2^L$};
\node at (2.8,2) {\small $4^L$};
\node at (3.2,4) {\small $1^R$};
\node at (3.2,2) {\small $3^R$};

\draw[step=0.4,thin,shift={(0.2,0)}] (8.799,1.999) grid (10.8,4);
\draw[ultra thick] (9,3.2) -- (8.6,3.2) -- (8.6,2.8) -- (9,2.8) -- (9,2) -- (9.8,2);
\draw[ultra thick] (9,3.2) -- (9,4) -- (9.8,4) -- (9.8,4.4) -- (10.2,4.4) -- (10.2,4);
\draw[ultra thick] (10.2,4) -- (11,4) -- (11,3.2) -- (11.4,3.2) -- (11.4,2.8) -- (11,2.8);
\draw[ultra thick] (9.8,2) -- (9.8,1.6) -- (10.2,1.6) -- (10.2,2) -- (11,2) -- (11,2.8);
\draw[ultra thick] (10.2,3.2) rectangle (10.6,3.6);

\node at (10.4,3.4) {\small $F$};
\node at (8.8,3)    {\small $L$};
\node at (11.2,3)   {\small $R$};
\node at (10,1.8)   {\small $B$};
\node at (10,4.2)   {\small $T$};

\node at (3,0.7) {\Large a)};
\node at (10,0.7) {\Large b)};
\end{tikzpicture}
\end{adjustbox}
\end{center}

\caption{a) A 4-room LMDP, with a terminal state $F$ and 8 other exit states; b) a single subtask with 5 terminal states $F,L,R,T,B$ that is equivalent to all 4 room subtasks. Rooms are numbered 1 through 4, left-to-right, then top-to-bottom, and exit state $1^B$ refers to the exit $B$ of room $1$, etc.}
\label{fig:ex}
\end{figure}

\paragraph{Example 1:} Figure~\ref{fig:ex}a) shows an example 4-room LMDP with a single terminal state marked $F$, separate from the room but reachable in one step from the highlighted location. The rooms are only connected via a single doorway; hence if we partition the states by room, the subtask corresponding to each room has two terminal states in other rooms, plus the terminal state $F$ for the top right room. The 9 exit states in $\cE$ are highlighted and correspond to states next to doorways, plus $F$. Figure~\ref{fig:ex}b) shows a single subtask that is equivalent to all four room subtasks, since dynamics is shared inside rooms and the set of terminal states is the union of those of the subtasks.
Hence the number of equivalent subtasks is $C=1$, the number of non-terminal and terminal states of subtasks is $K=25$ and $N=5$, respectively, and the number of exit states is $E=9$.

\subsection{Subtask Compositionality}

During learning, the value estimate $\hat{v}$ changes frequently, and it is inefficient to solve all subtasks after each change. Instead, our approach is to use compositionality to obtain solutions to the subtasks without learning. The idea is to introduce several base LMDPs for each subtask $\cL_j$ such that {\em any} reward function $\cJ_j$ can be expressed as a combination of the reward functions of the base LMDPs.

Given a subtask $\cL_j=\langle\cS_j,\cT_j,\cP_j,\cR_j,\cJ_j\rangle$ as defined above, assume that the set $\cT_j$ contains $n$ states, i.e.~$\cT_j=\{\tSi_1,\ldots,\tSi_n\}$. We define $n$ base LMDPs $\cL_j^1,\ldots,\cL_j^n$, where each base LMDP is given by $\cL_j^k=\langle\cS_j,\cT_j,\cP_j,\cR_j,\cJ_j^k\rangle$. Hence the base LMDPs only differ in the reward of terminal states.
Concretely, we define the exponentiated reward as $z_j^k(\tSi)=1$ if $\tSi=\tSi_k$, and $z_j^k(\tSi)=0$ otherwise.
This corresponds to an actual reward of $\cJ_j^k(\tSi)=0$ for $\tSi=\tSi_k$, and $\cJ_j^k(\tSi)=-\infty$ otherwise.

Even though the exit reward $\cJ_j^k(\tSi)$ equals negative infinity for terminal states different from $\tSi_k$, this does not cause computational issues in the exponentiated space, since the value $z_j^k(\tSi)=0$ is well-defined in \eqref{eq:matrixz} and \eqref{eq:pi}.
Moreover, there are two good reasons for defining the rewards in this way. The first is that the rewards form a convenient basis that allows us to express {\em any} value estimate on the terminal states in $\cT_j$ as a linear combination of $z_j^1,\ldots,z_j^n$.
The second is that a value estimate $\hat{z}(\tSi)=0$ can be used to {\em turn off} terminal state $\tSi$, since the definition of the optimal policy in \eqref{eq:pi} assigns probability $0$ to any transition that leads to a state $\tSi$ with $\hat{z}(\tSi)=0$. 
This is the reason that we do not need the sets of terminal states to be equal for equivalent subtasks.

Now assume that we solve the base LMDPs to obtain the optimal value functions $z_j^1,\ldots,z_j^n$. Also assume a given value estimate $\hat{v}$ for the terminal states in $\cT_j$, i.e.~$\cJ_j(\tSi)=\hat{v}(\tSi)$ for each $\tSi\in\cT_j$. Then we can write the exponentiated reward $\hat{z}(\tSi)=e^{\hat{v}(\tSi)/\lambda}$ of each terminal state as
\begin{equation}\label{eq:comp}
\hat{z}(\tSi) = \sum_{k=1}^n w_kz_j^k(\tSi) = \sum_{k=1}^n \hat{z}(\tSi_k) z_j^k(\tSi),
\end{equation}
where each weight is simply given by $w_k=\hat{z}(\tSi_k)$. This is because for a given terminal state $\tSi_\ell\in\cT_j$, the value $z_j^k(\tSi_\ell)$ equals $0$ for $k\neq \ell$, so the weighted sum simplifies to $w_\ell z_j^\ell(\tSi_\ell)=w_\ell\cdot 1=\hat{z}(\tSi_\ell)$.

Due to compositionality, we can now write the estimated value of each non-terminal state $s\in\cS_i$ as
\begin{align}\label{eq:comp3}
\hat{z}(s)=\sum_{k=1}^n \hat{z}(\tSi_k) z_j^k(s) \;\; \forall s\in\cS_i,\forall\cL_i\in\cC_j.
\end{align}
Here, the terminal states $\tSi_1,\ldots,\tSi_n$ are by definition exit states in $\cE$. If we have access to a value estimate $\hat{z}_\cE:\cE\rightarrow\mathbb{R}$ on exit states, as well as the value functions $z_j^1,\ldots,z_j^n$ of all base LMDPs, we can thus use \eqref{eq:comp3} to express the value estimate of each other state without learning. Hence~\eqref{eq:comp3} is a form of value function decomposition, allowing us to express the values of arbitrary states in $\cS$ in terms of value functions with smaller domains. Concretely, there are $O(CN)$ base LMDPs, each with $O(M)$ values, so in total we need $O(CMN+E)$ values for the decomposition.

\paragraph{Example 1:} 
In the 4-room example, there are five base LMDPs with value functions $z^F$, $z^L$, $z^R$, $z^T$ and $z^B$, respectively. Given an initial value estimate $\hat{z}_\cE$ for each exit state in $\cE$, a value estimate of any state in the top left room is given by $\hat{z}(s)=\hat{z}_\cE(1^B) z^B(s) + \hat{z}_\cE(1^R) z^R(s)$, where we use $\hat{z}_\cE(F)=\hat{z}_\cE(L)=\hat{z}_\cE(T)=0$ to indicate that the terminal states $F$, $L$ and $T$ are not present in the top left room. We need $CMN = 125$ values to store the value functions of the 5 base LMDPs, and $E=9$ values to store the value estimates of all exit states. Although this is more than the 100 states of the original LMDP, if we increase the number of rooms to $X\times Y$, the term $CMN$ is a constant as long as all rooms have equivalent dynamics, and the number of exit states is $E=(2X-1)(2Y-1)$, which is much smaller than the $25XY$ total states. For $10\times 10$ rooms, the value function decomposition requires $486$ values to represent the values of $2{,}500$ states.\\

The 4-room example is limited in the sense that changing the configuration and size of the rooms may break the assumption of equivalence, which in turn makes the hierarchical approach less powerful. However, the notion of equivalence is naturally associated with factored (L)MDPs, in which the state is factored into a set of variables $\cV=\{v_1,\ldots,v_m\}$, i.e.~$\cS=\cD(v_1)\times\cdots\times\cD(v_m)$, where $D(v_i)$ is the domain of variable $v_i$, $1\leq i\leq m$.
Concretely, if there is a subset of variables $\cU\subset\cV$ such that the transitions among $\cU$ are independent of the variables in $\cV\setminus\cU$, then it is natural to partition the states based on their assignment to the variables in $\cV\setminus\cU$. Consequently, there is a single equivalent subtask whose set of states is $\times_{v\in\cU}D(v)$, i.e.~all partial states on the variables in $\cU$.

\paragraph{Example 2:} The Taxi domain~\citep{dietterich2000hierarchical} is described by three variables: the location of the taxi ($v_1$), and the location and destination of the passenger ($v_2$ and $v_3$). Since the location of the taxi is independent of the other two, it is natural to partition the states according to the location and destination of the passenger. Each partition consists of the possible locations of the taxi, defining a unique equivalent subtask whose terminal states are the locations at which the taxi can pick up or drop off passengers. Since there are 16 valid combinations of passenger location and destination, there are 16 such equivalent subtasks.
\citet{dietterich2000hierarchical} calls this condition {\em max node irrelevance}, where ``max node'' refers to a given subtask.

\subsection{Eigenvector Approach}

If the dynamics $\cP$ and the state costs $\cR, \cJ$ are known, we can use the power method to solve the original LMDP $\cL$ by composing individual solutions of the subtask LMDPs $\cL_i$.
In this case, we define Bellman equations in \eqref{eq:matrixz} to solve the base LMDPs of all equivalence classes.
To compute the values of the original LMDP $\cL$ for the exit states in $\cE$, the compositionality relation in~\eqref{eq:comp3} provides us with an additional system of linear equations, one for each non-terminal exit state. We can reformulate this additional system of equations in matrix form defined for the exit states ${\bf z}_\cE$:
\begin{equation}\label{eq:exits}
{\bf z}_\cE=G{\bf z}_\cE.
\end{equation}
Here, the matrix $G$ contains the values of the base LMDPs according to~\eqref{eq:comp3}.
We can thus use the power method on this system of linear equations to obtain the values of all exit states in $\cE$.

\paragraph{Example 1:} In the 4-room example, the row in $G$ corresponding to $\hat{z}_\cE(2^L)$ contains the element $z^B(2^L)$ in the column for $\hat{z}_\cE(1^B)$, and the element $z^R(2^L)$ in the column for $\hat{z}_\cE(1^R)$, while all other elements equal $0$. While the flat approach requires one run of the power method on a large matrix, 
our hierachical approach needs five runs of the power method on significantly reduced 
 matrices (these runs can be parallelized), and one additional run on a $8\times 8$ matrix, corresponding to~\eqref{eq:exits}.\\ 

We remark that we do not explicitly represent the values of states in $\cS\setminus\cE$ since they are given by~\eqref{eq:comp3}.
Since we can now obtain the value $z(s)$ of each state $s\in\cS$, we can define the optimal policy directly in terms of the values $z$ and~\eqref{eq:pi}. Hence unlike most approaches to hierarchical reinforcement learning, the policy does not select among subtasks, but instead depends directly on the decomposed value estimates.

\subsection{Online and Intra-task Learning}


In the online learning case, we need to maintain estimates $\hat{z}_j^1,\ldots,\hat{z}_j^n$ of the value functions of the base LMDPs associated with each equivalent subtask $\cL_j$.
These estimates can be updated using the Z-learning rule~\eqref{eqn:zlearning-imp} after each transition.
But to make learning more efficient, we can use a single transition $(s_t,r_t,s_{t+1})$ with $s_t\in\cS_j$ to update the values of {\em all} base LMDPs associated with $\cL_j$ simultaneously. This is known in the literature as intra-task learning~\citep{Kaelbling93,conf/icaps/Jonsson16}.


Given the estimates $\hat{z}_j^1,\ldots,\hat{z}_j^n$, we could then formulate and solve the same system of linear equations in~\eqref{eq:comp3} to obtain the value estimates of exit states. However, it is impractical to solve this system of equations every time we update 
$\hat{z}_j^1,\ldots,\hat{z}_j^n$. Instead, we explicitly maintain estimates $\hat{z}_\cE$ of the values of exit states in the set $\cE$, and update these values incrementally. 
For that, we turn~\eqref{eq:comp3} into an update rule:
\begin{align}\label{eq:comprule}
\hat{z}_\cE(s) \leftarrow &(1 - \alpha_\ell) \hat{z}_\cE(s) + \alpha_\ell \sum_{k=1}^n \hat{z}_j^k(s) \hat{z}_\cE(\tSi_k).
\end{align}
The question is when to update the value of an exit state. We propose several alternatives:
\begin{itemize}[leftmargin=2em]
\item[$V_1$:] Update the value of an exit state $s\in\cE_i$ each time we take a transition from $s$.
\item[$V_2$:] When we reach a terminal state of the subtask $\cL_i$, update the values of all exit states in $\cE_i$.
\item[$V_3$:] When we reach a terminal state of the subtask $\cL_i$, update the values of all exit states in $\cE_i$ and all exit states of subtasks in the equivalence class $\cC_j$ of $\cL_i$.
\end{itemize}
Again, the estimated policy $\pi$ is defined directly by the value estimates $\hat{z}$ and~\eqref{eq:pi}, and thus does not select among subtasks. Below is the pseudo-code of the proposed algorithm.

\begin{algorithm}[H]
\setstretch{1.12}
\renewcommand{\thealgorithm}{}
\caption{Online and Intra-Task Learning Algorithm}
\begin{algorithmic}[1]
\STATE {\bf Input:} An LMDP ${\cL = \langle \cS, \cT, \cP, \cR, \cJ \rangle}$ and
a partition $\{\cS_i\}_{i=1}^L$ of $\cS$ \newline
A set $\{\cC_1,\ldots,\cC_C\}$ of equivalent subtasks and related base LMDPs $\mathcal{L}_j^k = \langle \cS_j, \cT_j, \cP_j, \cR_j, \cJ_j^k \rangle$

\STATE {\bf Initialization:} \newline
$\hat z_\cE(s) := 1 \;\; (\forall s \in \cE)$ \COMMENT {high-level Z function approximation} \newline
$\hat z_j^k(s) := 1$ \COMMENT{base LMDPs $1 \dots |\cT_j|$ for each equivalent subtask $\cL_j$}

\WHILE{$\text{termination condition is not met}$}
\STATE observe transition $s_t, r_t, s_{t+1} \sim \hat\pi(\cdot|s_{t})$, where $s_t\in\cS_i$ and $\cL_i\in\cC_j$
\STATE $\text{update lower-level estimations } \hat z_j^k(s_t)$ \text{using} \eqref{eqn:zlearning-imp}
\IF[$s_t$ is an exit or $s_{t+1}$ is terminal for current subtask $\cL_j$]{$s_t\in\cE$ or $s_{t+1} \in \cT_j$}
\STATE $\text{apply \eqref{eq:comprule} to update $\hat z_\cE$ using variant } V_1$, $V_2$ or $V_3$
\ENDIF
\ENDWHILE

\end{algorithmic}
\end{algorithm}

\begin{figure*}[!htb]
\setlength{\belowcaptionskip}{-10pt}
\centering
\includegraphics[scale=0.18]{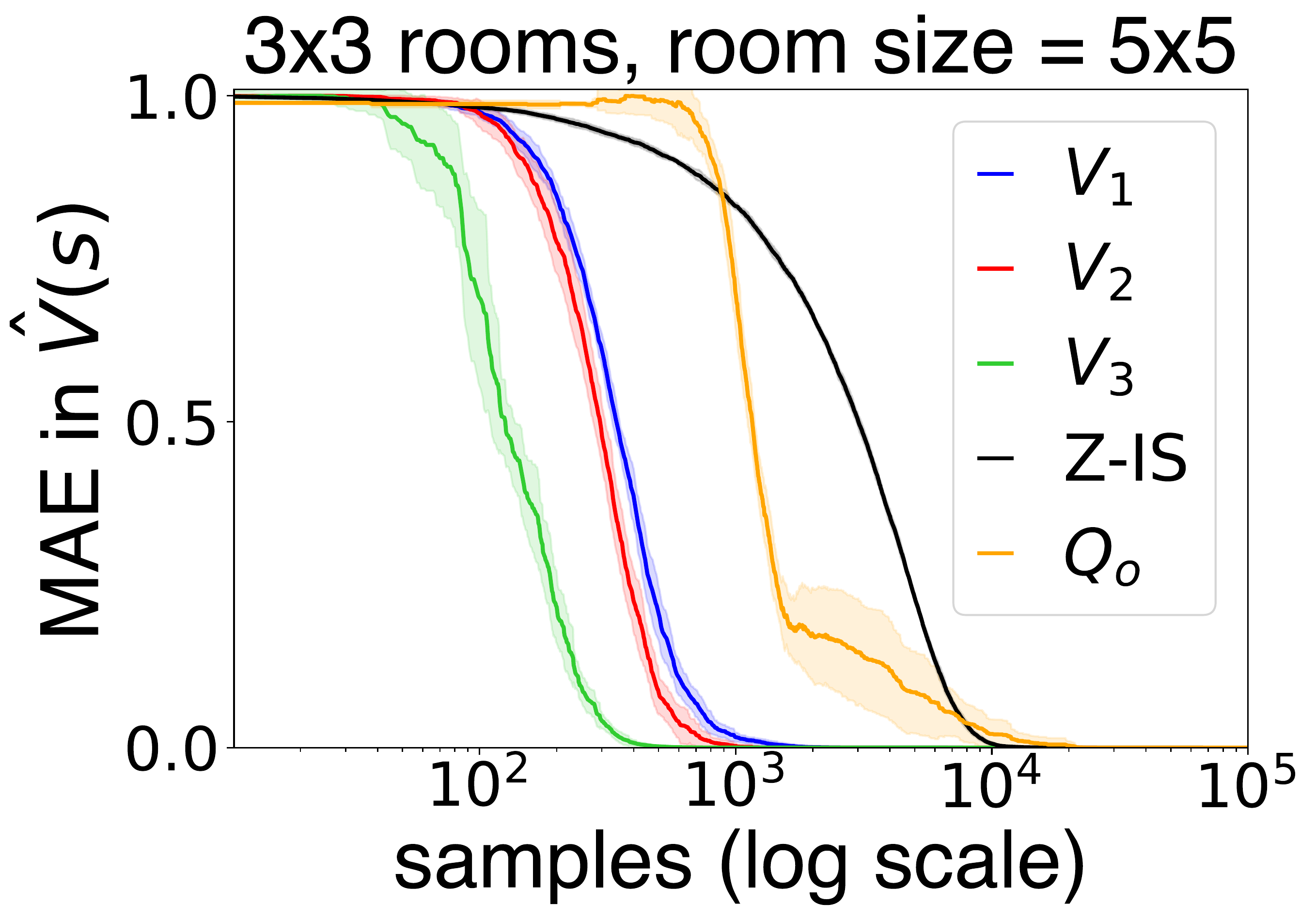}
\includegraphics[scale=0.18]{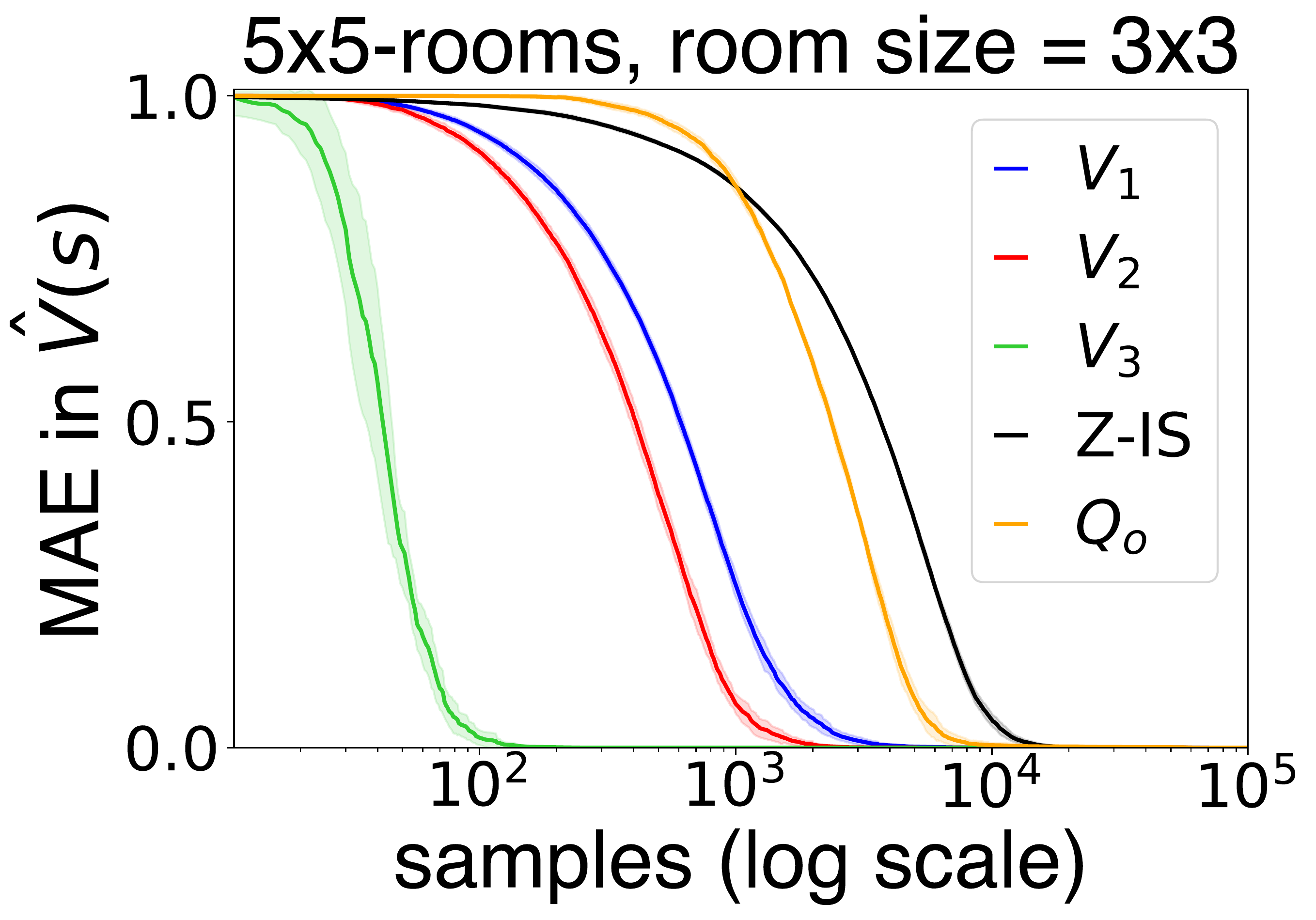}
\includegraphics[scale=0.18]{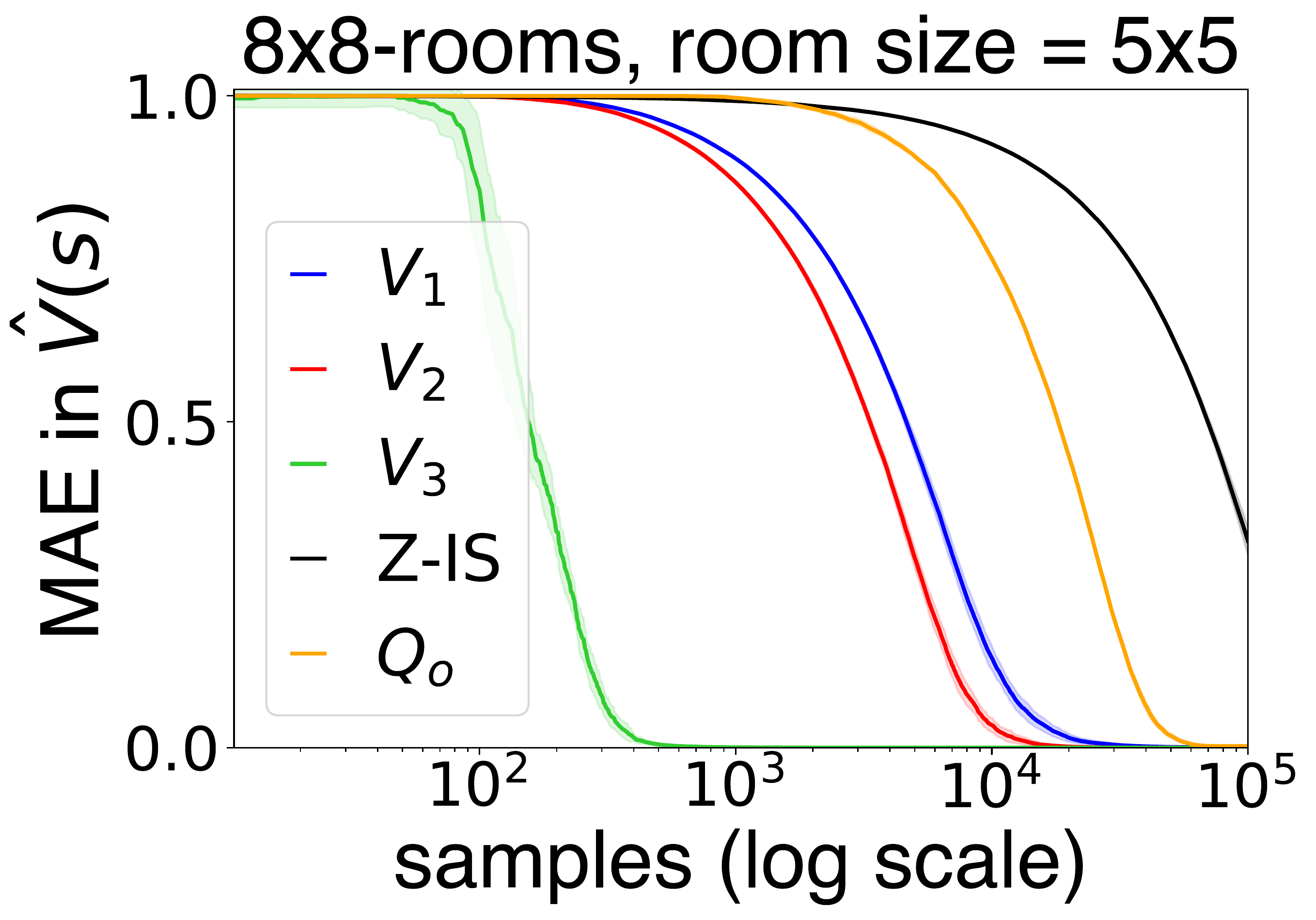}\\
\caption{ Results for $3\times 3$ rooms of size $5 \times 5$ (left);
$5\times 5$ rooms of size $3 \times 3$ (center); $8 \times 8$ rooms of size $5\times 5$ (right).}
\label{fig:errors_grid}
\end{figure*}

\subsection{Analysis}
Let $\cL=\langle\cS,\cT,\cP,\cR,\cJ\rangle$ be an LMDP, and let $\cL_i=\langle\cS_i,\cT_i,\cP_i,\cR_i,\cJ_i\rangle$ be a subtask associated with the partition $\cS_i\subseteq\cS$. Let $z$ denote the optimal value of $\cL$, and let $z_i$ denote the optimal value of $\cL_i$.

\begin{lemma}\label{lemma:same}
If the reward of each terminal state $\tSi\in\cT_i$ equals its optimal value in $\cL$, i.e.~$z_i(\tSi)=z(\tSi)$, the optimal value of each non-terminal state $s\in\cS_i$ equals its optimal value in $\cL$, i.e.~$z_i(s)=z(s)$.
\end{lemma}

\begin{proof}
Since $\cP_i$ and $\cR_i$ are the restriction of $\cP$ and $\cR$ onto $\cS_i$, for each $s\in\cS_i$ we have
\begin{align*}
    z_i(s) &= e^{\cR_i(s)/\lambda} \sum_{s'}\cP_i(s'|s)z_i(s') \\
    &= e^{\cR(s)/\lambda} \sum_{s'}\cP(s'|s)z_i(s'),
\end{align*}
which is 
 the same Bellman equation as for $z(s)$.
Since $z_i(\tSi)=z(\tSi)$ for each terminal state $\tSi\in\cT_i$, we immediately obtain $z_i(s)=z(s)$ for each non-terminal state $s\in\cS_i$.
\end{proof}
As an consequence of Lemma~\ref{lemma:same}, assigning the optimal value $z(\tSi)$ to each exit state $\tSi\in\cE$ yields a solution to~\eqref{eq:exits}, which is thus guaranteed to have a solution with eigenvalue~$1$. Lemma~\ref{lemma:same} also guarantees that we can use \eqref{eq:comp3} to compute the optimal value of any arbitrary state given optimal values of the base LMDPs and the exit states. The only necessary conditions needed for convergence to the optimal value function is that $(i)$ $\{\cS_i\}_{i=1}^L$ is a proper partition of the state space; and $(ii)$ the set of terminal states $\cT_i$ of each subtask $\cL_i$ includes all states reachable in one step from $\cS_i$.

\begin{lemma}
The solution to \eqref{eq:exits} is unique.
\end{lemma}

\begin{proof}
By contradiction. Assume that there exists a solution ${\bf z}_\cE'$ which is different from the optimal values ${\bf z}_\cE$. We can extend ${\bf z}$ and ${\bf z}'$ to all states in $\cS$ by applying \eqref{eq:comp3}. Due to the same argument as in the proof of Lemma~\ref{lemma:same}, the solution ${\bf z}'$ satisfies the Bellman optimality equation of all states in $\cS$. Hence ${\bf z}'$ is an optimal value function for the original LMDP $\cL$, which contradicts that ${\bf z}'$ is different from ${\bf z}$ since the Bellman optimality equations have a unique solution.
\end{proof}

\begin{lemma}
For each subtask $\cL_i$ and state $s\in\cS_i^+$, it holds that $z_i^1(s)+\cdots+z_i^n(s)\leq 1$.
\end{lemma}

\begin{proof}
By induction. The base case is given by terminal states $t_\ell\in\cT_i$, in which case $z_i^1(t_\ell)+\cdots+z_i^n(t_\ell) = z_i^\ell(t_\ell) = 1$. For $s\in\cS_i$, the Bellman equation for each base LMDP yields
\begin{align*}
    \sum_{k=1}^n z_i^k(s)=e^{\cR_i(s)/\lambda}\sum_{s'}\cP(s'|s)\sum_{k=1}^n z_i^k(s').
\end{align*}

Since $\cR_i(s)=\cR(s)<0$ holds by assumption, and since $z_i^1(s')+\cdots+z_i^n(s')\leq 1$ holds for each $s'$ by hypothesis of induction, it follows that $z_i^1(s)+\cdots+z_i^n(s)\leq 1$.
\end{proof}
As a consequence, just like the matrix $RP$ in \eqref{eq:matrixz}, the matrix $G$ in \eqref{eq:exits} has spectral radius at most $1$, and hence the power method is guaranteed to converge to the unique solution with largest eigenvalue $1$, corresponding to the optimal values of the exit states.

The convergence rate of the power method is exponential in $\gamma<1$, the eigenvalue of $RP$ or $G$ with second largest value and independent of the state space.
The average running time scales linearly with the number of non-zero elements in $RP$ or $G$~\cite{TodorovNIPS2007}, which is drastically reduced compared to the non-hierarchical approach. 
More precisely, given an upper bound $B$ on the support of $\cP$ and a sparse representation, the matrix multiplication in~\eqref{eq:matrixz} has complexity $\mathcal{O}(BS)$. In comparison, the matrix multiplication of the $\mathcal{O}(CN)$ base LMDPs has complexity $\mathcal{O}(BK)$, while the matrix multiplication in \eqref{eq:exits} has complexity $\mathcal{O}(NE)$. Hence the hierarchical approach is competitive whenever $\mathcal{O}(CNBK+NE)$ is smaller than $\mathcal{O}(BS)$. In a $10\times 10$ room example, $CNBK+NE=500+1{,}805=2{,}305$, while $BS=10{,}000$.


\section*{Experiments}
\begin{center}
\begin{figure}[!h]
\setlength{\belowcaptionskip}{-25pt}
\includegraphics[scale=0.18]{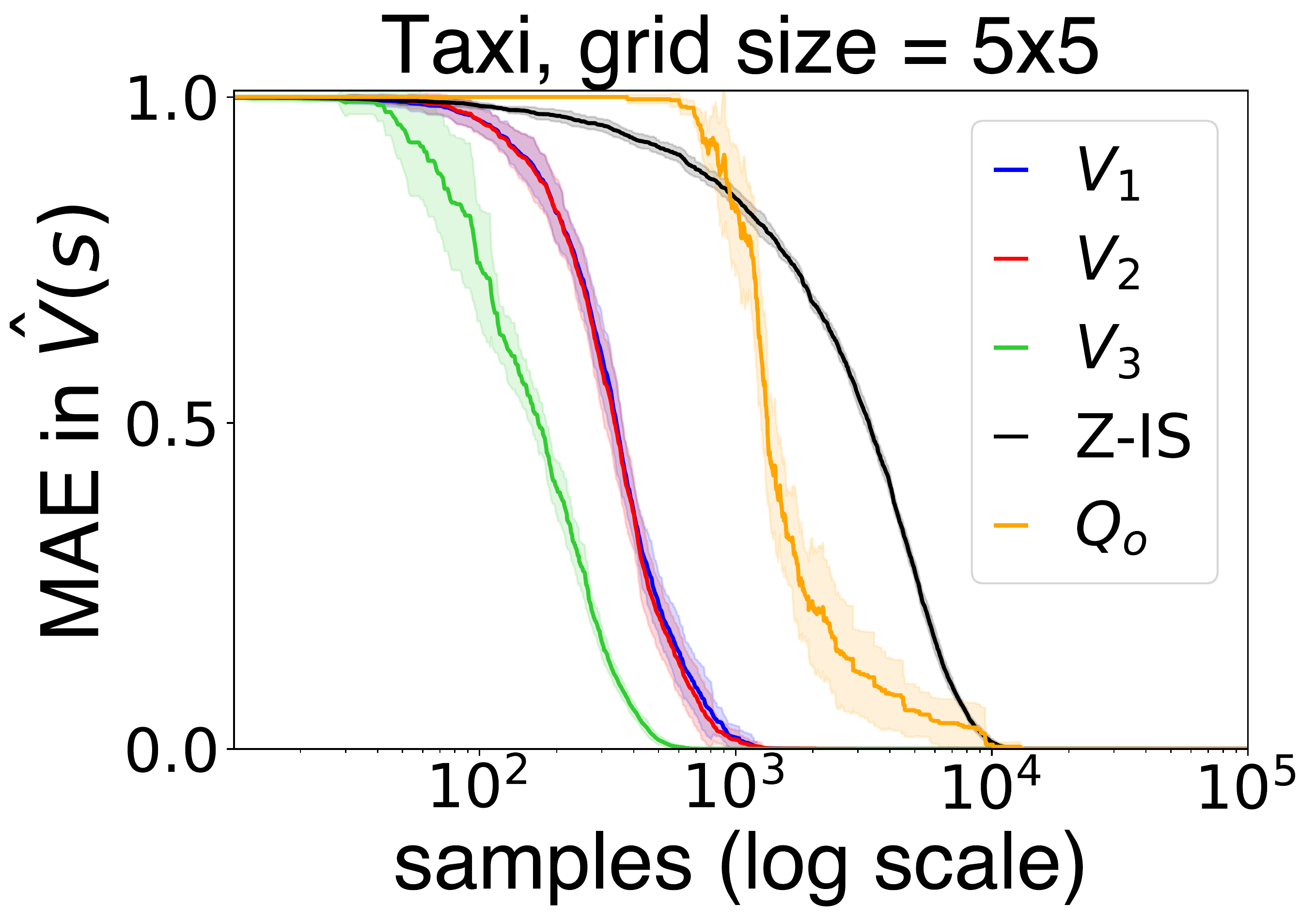}
\includegraphics[scale=0.18]{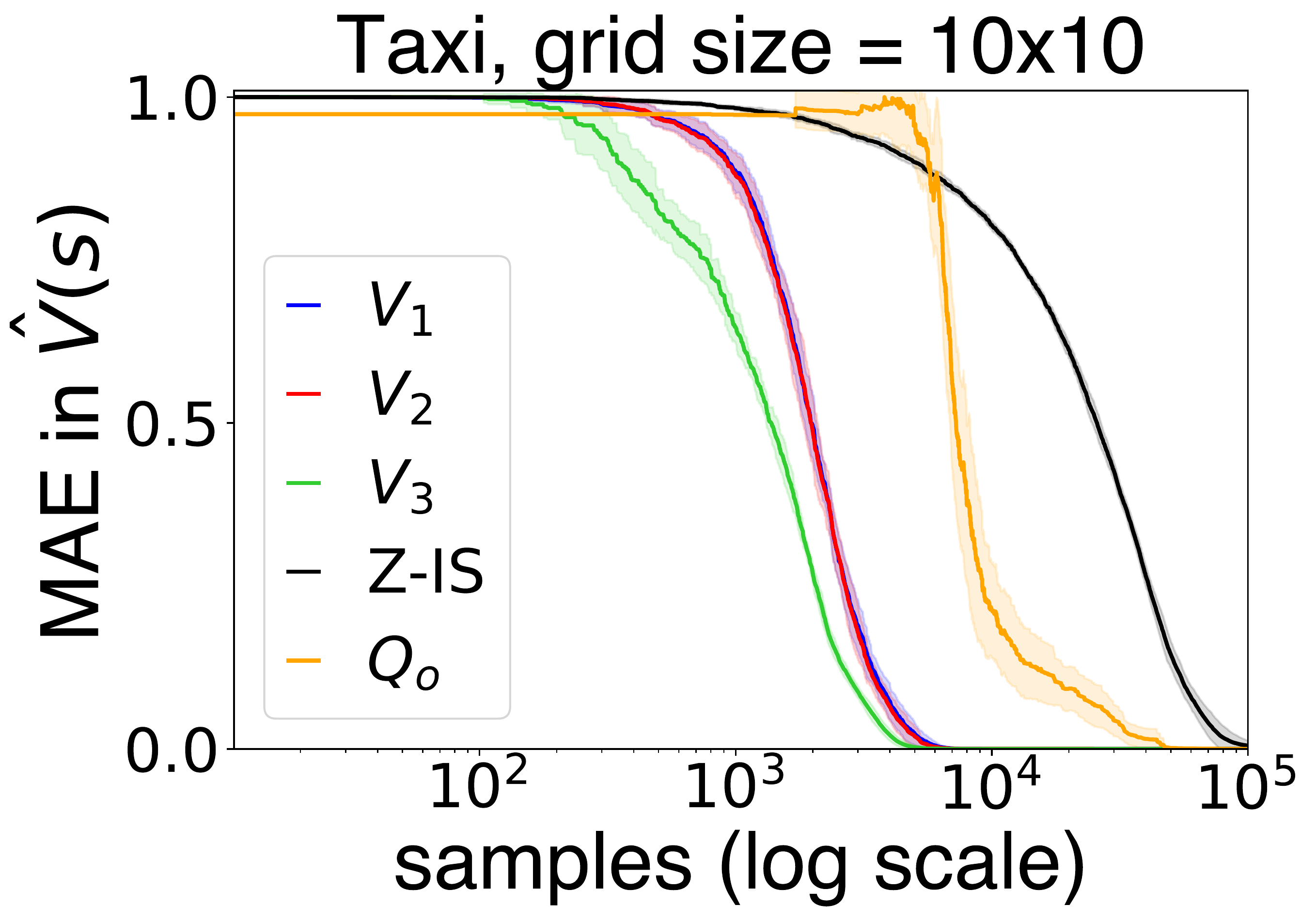}
\caption{Results for Taxi for $5 \times 5$ and $10 \times 10$ (resp.) grids.}
\label{fig:errors_taxi}
\end{figure}
\end{center}

We now evaluate the proposed learning algorithm in the two previous examples.\footnote{Code available at https://github.com/guillermoim/HRL\_LMDP}
The objective of this evaluation is to analyze empirically the different update alternatives ($V_1$, $V_2$, and $V_3$), and
to compare against a flat approach which exploits the benefits of LMDPs without the hierarchy (Z-IS), and the hierarchical approach based on options ($Q_o$)~\cite{sutton1999between}. Our main objective is to empirically show that our approach is more sample efficient than the other algorithms. We run each algorithm with four different random seeds to analyze the average MAE (mean absolute error) against the optimal value function (computed separately) and its standard deviation over the number of samples. Since the value functions are different for Q-learning and LMDP methods, we present the self-normalized MAE (Figures \ref{fig:errors_grid} and \ref{fig:errors_taxi}) for different configurations and domains. Further, for a fair comparison between approaches, we only use the exit set for calculating the MAE.


In all experiments, the learning rates for each abstraction level is $\alpha_\ell(t) = c_\ell / (c_\ell + n)$ where $n$ represents the episode each sample $t$ belongs to. We empirically optimize the constant $c_\ell$ for each domain. For LMDPs, we use a temperature $\lambda=1$, which provides good results. $Q_o$ solves an equivalent MDP with {\em deterministic} actions, which should actually give it an advantage. For fairness, $Q_o$ obtains the same per-step negative reward, exploits the same equivalence classes, learns the same subtasks (i.e.~reach a terminal state), and has knowledge of which options are available in each state.

\textbf{Rooms Domain.}
We analyze the performance for different room sizes and number of rooms (Figure \ref{fig:errors_grid}). In all configurations the proposed hierarchical approach outperfoms Z-IS and $Q_o$. Concretely, $Q_o$ suffers from non-stationarity: initial option executions will incur more negative reward than later executions, which causes high-level Q-learning updates to be {\em incorrect}, and it takes the learner significant time to recover from this.

Figure~\ref{fig:errors_grid} (left) shows results for $3\times 3$ rooms of size $5\times 5$ and Figure~\ref{fig:errors_grid} (center) shows results for $5\times 5$ rooms of size $3\times 3$. Both scenarios have $225$ interior states.
The difference between variants $V_1$, $V_2$ and $V_3$ is more pronounced in the second case, when the number of subtasks increases (more rooms) and the partition for each subtask is smaller (smaller rooms). Figure~\ref{fig:errors_grid} (right) shows how the method scales with the number of rooms of size $5\times 5$.
Again, variant $V_3$ has the best performance, in this case by a larger 
margin than before.





\textbf{Taxi Domain.}
To allow comparison between all the methods, we adapted the Taxi domain as follows: when the taxi is at the correct pickup location, it can transition to a state with the passenger in the taxi.
In a wrong pickup location, it can instead transition to a terminal state with large negative reward (simulating an unsuccessful pick-up).
When the passenger is in the taxi, it can be dropped off at any pickup location, successfully completing the task whenever dropped at the correct destination.

Figure~\ref{fig:errors_taxi} shows results in two instances of size $5 \times 5$ ($408$ states) and $10 \times 10$ ($1608$ states). 
Again, the proposed hierarchical approach outperforms Z-IS and $Q_o$.
In this case, the difference between $V_1$, $V_2$ and $V_3$ is less pronounced, even when the grid size increases. One possible explanation is the small number of exit states in this problem.


\section*{Discussion and Conclusion}

In this paper we have introduced a novel approach to hierarchical reinforcement learning that focuses on the class of linearly-solvable Markov decision processes.
Using subtask compositionality, we can decompose the value function and derive algorithms that converge to the optimal value function.
To the best of our knowledge, our approach is the first to exploit both the concurrent compositionality enabled by LMDPs together with hierarchies and intra-task learning to obtain globally optimal policies efficiently.

The proposed hierarchical decomposition leads to a new form of zero-shot learning that allows to incorporate subtasks that belong to an existing equivalent class without additional learning effort.
For example, adding new rooms in our example.
This is in contrast with existing methods that only exploit linear compositionality of tasks.

Our approach is limited to OR compositionality of subtasks, but there is no fundamental limitation that prevents arbitrary compositions.
The benefits of hierarchies can be combined for example, with the extended value functions proposed in~\citet{algebra}.




\bibliography{aaai22}

\begin{thebibliography}{29}
\providecommand{\natexlab}[1]{#1}

\bibitem[{Barreto et~al.(2019)Barreto, Borsa, Hou, Comanici, Ayg\"{u}n, Hamel,
  Toyama, Hunt, Mourad, Silver, and Precup}]{optionkeyb}
Barreto, A.; Borsa, D.; Hou, S.; Comanici, G.; Ayg\"{u}n, E.; Hamel, P.;
  Toyama, D.; Hunt, J.; Mourad, S.; Silver, D.; and Precup, D. 2019.
\newblock {The Option Keyboard: Combining Skills in Reinforcement Learning}.
\newblock In \emph{Advances in Neural Information Processing Systems 32},
  13031--13041.

\bibitem[{Barto and Mahadevan(2003)}]{bartomahadevan}
Barto, A.~G.; and Mahadevan, S. 2003.
\newblock Recent Advances in Hierarchical Reinforcement Learning.
\newblock \emph{Discrete Event Dynamic Systems}, 13(1–2): 41–77.

\bibitem[{Bas-Serrano et~al.(2021)Bas-Serrano, Curi, Krause, and
  Neu}]{pmlr-v130-bas-serrano21a}
Bas-Serrano, J.; Curi, S.; Krause, A.; and Neu, G. 2021.
\newblock { Logistic Q-Learning }.
\newblock In \emph{Proceedings of The 24th International Conference on
  Artificial Intelligence and Statistics}, volume 130, 3610--3618. PMLR.

\bibitem[{Boutilier, Dearden, and Goldszmidt(1995)}]{boutilier95}
Boutilier, C.; Dearden, R.; and Goldszmidt, M. 1995.
\newblock Exploiting Structure in Policy Construction.
\newblock In \emph{Proceedings of The 14th International Joint Conference on
  Artificial Intelligence}.

\bibitem[{da~Silva, Durand, and Popovi\'{c}(2009)}]{animation}
da~Silva, M.; Durand, F.; and Popovi\'{c}, J. 2009.
\newblock Linear Bellman Combination for Control of Character Animation.
\newblock \emph{ACM Trans. Graph.}, 28(3).

\bibitem[{Dietterich(2000)}]{dietterich2000hierarchical}
Dietterich, T.~G. 2000.
\newblock Hierarchical reinforcement learning with the {MAXQ} value function
  decomposition.
\newblock \emph{J. Artif. Intell. Res.}, 13: 227--303.

\bibitem[{Dvijotham and Todorov(2013)}]{djbook}
Dvijotham, K.; and Todorov, E. 2013.
\newblock Linearly Solvable Optimal Control.
\newblock In Lewis, F.~L.; and Liu, D., eds., \emph{Reinforcement Learning and
  Approximate Dynamic Programming for Feedback Control}, chapter~6, 119--141.
  John Wiley \& Sons.

\bibitem[{Haarnoja et~al.(2018{\natexlab{a}})Haarnoja, Pong, Zhou, Dalal,
  Abbeel, and Levine}]{haarnoja2018composable}
Haarnoja, T.; Pong, V.; Zhou, A.; Dalal, M.; Abbeel, P.; and Levine, S.
  2018{\natexlab{a}}.
\newblock Composable deep reinforcement learning for robotic manipulation.
\newblock In \emph{2018 IEEE international conference on robotics and
  automation (ICRA)}, 6244--6251. IEEE.

\bibitem[{Haarnoja et~al.(2018{\natexlab{b}})Haarnoja, Zhou, Abbeel, and
  Levine}]{pmlr-v80-haarnoja18b}
Haarnoja, T.; Zhou, A.; Abbeel, P.; and Levine, S. 2018{\natexlab{b}}.
\newblock {Soft Actor-Critic: Off-Policy Maximum Entropy Deep Reinforcement
  Learning with a Stochastic Actor}.
\newblock In \emph{Proceedings of the 35th International Conference on Machine
  Learning}, volume~80, 1861--1870. PMLR.

\bibitem[{Hunt et~al.(2019)Hunt, Barreto, Lillicrap, and
  Heess}]{hunt2019composing}
Hunt, J.; Barreto, A.; Lillicrap, T.; and Heess, N. 2019.
\newblock Composing entropic policies using divergence correction.
\newblock In \emph{International Conference on Machine Learning}, 2911--2920.
  PMLR.

\bibitem[{Jonsson and G{\'o}mez(2016)}]{conf/icaps/Jonsson16}
Jonsson, A.; and G{\'o}mez, V. 2016.
\newblock {Hierarchical Linearly-Solvable Markov Decision Problems}.
\newblock In \emph{Proceedings of the 26th International Conference on
  Automated Planning and Scheduling (ICAPS)}.

\bibitem[{Kaelbling(1993)}]{Kaelbling93}
Kaelbling, L.~P. 1993.
\newblock {Learning to Achieve Goals}.
\newblock In \emph{Proceedings of the International Joint Conference on
  Artificial Intelligence (IJCAI)}, 1094--1099.

\bibitem[{Kappen(2005)}]{Kappen2005}
Kappen, H.~J. 2005.
\newblock Linear Theory for Control of Nonlinear Stochastic Systems.
\newblock \emph{Phys. Rev. Lett.}, 95: 200--201.

\bibitem[{Kappen(2013)}]{kappenbook}
Kappen, H.~J. 2013.
\newblock Optimal control theory and the linear {Bellman} equation.
\newblock In D.~Barber, S.~C., A.~Taylan, ed., \emph{Bayesian Time Series
  Models}, chapter~17, 363--387. Cambridge University Press.

\bibitem[{Kappen, G\'omez, and Opper(2012)}]{KappenML2012}
Kappen, H.~J.; G\'omez, V.; and Opper, M. 2012.
\newblock Optimal control as a graphical model inference problem.
\newblock \emph{Machine Learning}, 87(2): 159--182.

\bibitem[{Koller and Parr(2000)}]{DBLP:conf/uai/KollerP00}
Koller, D.; and Parr, R. 2000.
\newblock Policy Iteration for Factored MDPs.
\newblock In \emph{Proceedings of the 16th Conference in Uncertainty in
  Artificial Intelligence}, 326--334.

\bibitem[{Kolobov, Mausam, and Weld(2012)}]{Kolobov2012}
Kolobov, A.; Mausam; and Weld, D.~S. 2012.
\newblock Discovering hidden structure in factored MDPs.
\newblock \emph{Artificial Intelligence}, 189: 19--47.

\bibitem[{Levine(2018)}]{levine2018reinforcement}
Levine, S. 2018.
\newblock Reinforcement learning and control as probabilistic inference:
  Tutorial and review.
\newblock \emph{arXiv preprint arXiv:1805.00909}.

\bibitem[{Mnih et~al.(2016)Mnih, Badia, Mirza, Graves, Lillicrap, Harley,
  Silver, and Kavukcuoglu}]{pmlr-v48-mniha16}
Mnih, V.; Badia, A.~P.; Mirza, M.; Graves, A.; Lillicrap, T.; Harley, T.;
  Silver, D.; and Kavukcuoglu, K. 2016.
\newblock Asynchronous Methods for Deep Reinforcement Learning.
\newblock In \emph{Proceedings of The 33rd International Conference on Machine
  Learning}, volume~48, 1928--1937. PMLR.

\bibitem[{Nangue~Tasse, James, and Rosman(2020)}]{algebra}
Nangue~Tasse, G.; James, S.; and Rosman, B. 2020.
\newblock A Boolean Task Algebra for Reinforcement Learning.
\newblock In \emph{Advances in Neural Information Processing Systems},
  volume~33, 9497--9507.

\bibitem[{Saxe, Earle, and Rosman(2017)}]{saxe2017hierarchy}
Saxe, A.~M.; Earle, A.~C.; and Rosman, B. 2017.
\newblock {Hierarchy through composition with multitask LMDPs}.
\newblock In \emph{International Conference on Machine Learning}, 3017--3026.
  PMLR.

\bibitem[{Strehl, Diuk, and Littman(2007)}]{DBLP:conf/aaai/StrehlDL07}
Strehl, A.~L.; Diuk, C.; and Littman, M.~L. 2007.
\newblock Efficient Structure Learning in Factored-State MDPs.
\newblock In \emph{Proceedings of the Twenty-Second {AAAI} Conference on
  Artificial Intelligence}, 645--650.

\bibitem[{Sutton, Precup, and Singh(1999)}]{sutton1999between}
Sutton, R.~S.; Precup, D.; and Singh, S. 1999.
\newblock Between {MDPs} and semi-{MDPs}: A framework for temporal abstraction
  in reinforcement learning.
\newblock \emph{Artificial intelligence}, 112(1): 181--211.

\bibitem[{Todorov(2006)}]{TodorovNIPS2007}
Todorov, E. 2006.
\newblock Linearly-solvable {M}arkov decision problems.
\newblock \emph{Advances in Neural Information Processing Systems (NIPS)},
  1369--1376.

\bibitem[{Todorov(2009)}]{TodorovNIPS2009}
Todorov, E. 2009.
\newblock Compositionality of optimal control laws.
\newblock \emph{Advances in Neural Information Processing Systems (NIPS)},
  1856--1864.

\bibitem[{Van~Niekerk et~al.(2019)Van~Niekerk, James, Earle, and
  Rosman}]{van2019composing}
Van~Niekerk, B.; James, S.; Earle, A.; and Rosman, B. 2019.
\newblock Composing value functions in reinforcement learning.
\newblock In \emph{International Conference on Machine Learning}, 6401--6409.
  PMLR.

\bibitem[{Vieillard, Pietquin, and Geist(2020)}]{NEURIPS2020_2c6a0bae}
Vieillard, N.; Pietquin, O.; and Geist, M. 2020.
\newblock Munchausen Reinforcement Learning.
\newblock In \emph{Advances in Neural Information Processing Systems},
  volume~33, 4235--4246.

\bibitem[{Wen et~al.(2020)Wen, Precup, Ibrahimi, Barreto, Van~Roy, and
  Singh}]{conf/nips/Wen20}
Wen, Z.; Precup, D.; Ibrahimi, M.; Barreto, A.; Van~Roy, B.; and Singh, S.
  2020.
\newblock {On Efficiency in Hierarchical Reinforcement Learning}.
\newblock In \emph{Proceedings of the 34th Conference on Neural Information
  Processing Systems (NeurIPS)}.

\bibitem[{Ziebart(2010)}]{ziebart}
Ziebart, B.~D. 2010.
\newblock \emph{Modeling Purposeful Adaptive Behavior with the Principle of
  Maximum Causal Entropy}.
\newblock Ph.D. thesis, USA.

\end{thebibliography}

\end{document}